\documentclass{article}

\usepackage{microtype}
\usepackage{graphicx}
\usepackage{subcaption}
\usepackage{booktabs} 

\usepackage{tcolorbox}
\tcbuselibrary{breakable}
\usepackage{enumitem}

\usepackage[
    colorlinks=true,
    linkcolor=blue,
    urlcolor=blue,
    citecolor=blue
]{hyperref}

\usepackage[accepted]{icml2026}

\usepackage{amsmath}
\usepackage{amssymb}
\usepackage{mathtools}
\usepackage{amsthm}
\usepackage{multirow}

\usepackage[capitalize,noabbrev]{cleveref}

\theoremstyle{plain}

\theoremstyle{definition}

\theoremstyle{remark}

\usepackage[textsize=tiny]{todonotes}

\icmltitlerunning{SafeRun: Enabling Determinism in LLM Planning for Running}

\begin{document}

\twocolumn[
  \icmltitle{SafeRun: Enabling Determinism in LLM Planning for Running}



  \icmlsetsymbol{equal}{*}
  \icmlsetsymbol{corresponding}{$\dagger$} 

  \begin{icmlauthorlist}
    \icmlauthor{Meilin Chen}{equal,xhs,corresponding}
    \icmlauthor{Zepeng Zhai}{equal,xhs}
    \icmlauthor{Jiaxuan Zhao}{xhs,iie}
    \icmlauthor{Yuan Lu}{xhs}

  \end{icmlauthorlist}

  \icmlaffiliation{xhs}{Xiaohongshu Inc.}
  \icmlaffiliation{iie}{Institute of Information Engineering, Chinese Academy of Sciences}
  \icmlcorrespondingauthor{Meilin Chen}{merlinis@zju.edu.cn}
  
  \icmlkeywords{Machine Learning, ICML}

  \vskip 0.3in
]



\printAffiliationsAndNotice{}  

\begin{abstract}
 Large Language Models enable flexible natural-language planning but remain unreliable in determinism-critical domains due to their probabilistic nature. This limitation is especially problematic in running planning, where violating safety rules can lead to safety risks. We propose SafeRun, a framework for deterministic LLM-based planning via a decoupled architecture. SafeRun separates soft interpretation by an LLM from hard constraint enforcement by a deterministic solver, ensuring strict safety constraints while preserving natural-language flexibility. To validate SafeRun, we build a comprehensive benchmark for running planning under realistic physiological and safety constraints. Experiments across five LLMs show that SafeRun achieves 100\% safety score (vs.\ 79.1\% PE average and 97.6\% CodeAct average) while maintaining competitive instruction-following scores. The SafeRun benchmark is publicly available at \href{https://huggingface.co/datasets/zzp-seeker/SafeRun-RunPlanning-Benchmark}{{huggingface}}.
\end{abstract}

\section{Introduction}

Recent advancements in Large Language Models \cite{gpt5, gpt4.1, deepseek-v3.2, qwen3.5} have significantly driven the development of autonomous agents capable of complex reasoning and task planning. However, a fundamental barrier limits the deployment of LLMs in determinism-critical domains: the inherent unpredictability of probabilistic models, as evidenced by prior evaluation studies \cite{chen2025unbiased}. At their core, LLMs are token-by-token generators rather than formal reasoning systems equipped with strict execution capabilities. 
This limitation is particularly evident in a typical determinism-critical domain, running planning, where training loads must be carefully controlled within strict physiological limits. In such scenario where overrunning can lead to severe physical injuries, this lack of determinism makes purely LLM-driven planning unacceptably risky.

\begin{figure}[t]
\centering
\includegraphics[width=\columnwidth]{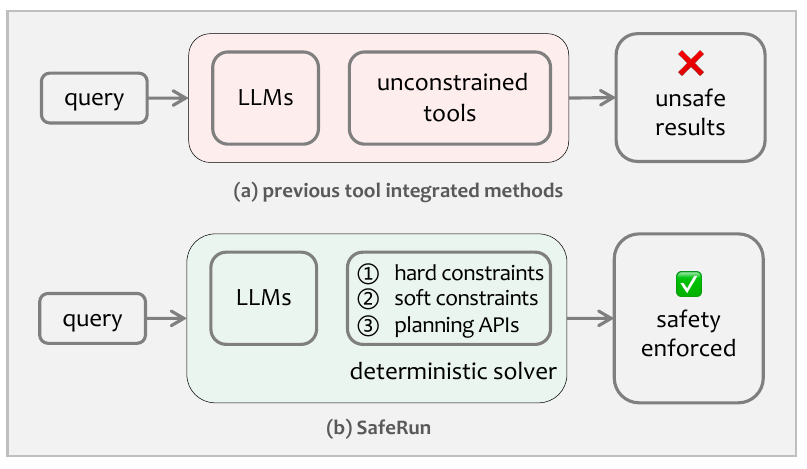}
\caption{Two paradigms for LLM-based planning in determinism-critical domains. (a)~\textbf{previous tool integrated methods}: the LLM uses external unconstrained tools (e.g., arbitrary codes), which fails to provide deterministic guarantees. (b)~\textbf{SafeRun (ours)}: the LLM is restricted to invoking a predefined deterministic solver, which is equipped with explicit constraint mechanisms, including hard constraints (e.g., safety rules), soft constraints (e.g., diversity and balance objectives), and predefined planning APIs.}
\vspace{-1.5em}
\label{fig:motivation}
\end{figure}

As illustrated in Figure~\ref{fig:motivation}, previous pure LLM based and tool integrated approaches ~\cite{hao2025large, shen2024hugginggpt, yao2024tree, zhao2024large, besta2024graph, hao2023reasoning} (refer to Appendix~\ref{Related Works} for detailed analysis of related works) mainly focus on complex-constraint planning problems, such as travel planning, by either relying solely on the reasoning capabilities of LLMs or augmenting them with external unconstrained tools. While these planning algorithms have shown promising results, they struggle to bridge the gap between probabilistic generation and the need for absolute reliability in determinism-critical settings. 

To address this challenge, as shown in Figure~\ref{fig:motivation}, we propose SafeRun, a novel framework that enables deterministic LLM-based planning for running. Unlike prior paradigms~\cite{hao2025large} that instruct the LLM to write arbitrary planning codes, SafeRun restricts the LLM to invoking a predefined deterministic solver, which is equipped with explicit constraint mechanisms, including hard constraints (e.g., safety rules), soft constraints (e.g., diversity and balance objectives), and predefined planning APIs.

Our key insight is that determinism-critical planning tasks can be decomposed into two distinct phases: (1) soft interpretation — understanding user intent, and determining how to invoke predefined solver; and (2) hard enforcement — ensuring that all safety constraints are strictly satisfied through a predefined formal deterministic solver. 
In this design, the LLM no longer directly constructs executable plans. Instead, it serves as a natural-language interface that translates user requests into solver-compatible specifications. 
By assigning the first phase to LLMs and the second to deterministic systems, SafeRun achieves both the flexibility of natural language interaction and the reliability required for determinism-critical planning.

To evaluate SafeRun under realistic and diverse conditions, we construct a benchmark through a structured pipeline curated by 10 human experts (bachelor’s degree or higher), consisting of 100 high-quality multi-week running planning queries and 10 diverse runner profiles covering a wide range of training goals and physiological constraints. By systematically pairing each query with every profile, we obtain 400 profile--query combinations for evaluating both natural-language understanding and safety-constrained planning.

We evaluate several state-of-the-art LLMs, including both proprietary and open-source models such as GPT-5~\citep{gpt5}, GPT-4.1~\citep{gpt4.1}, GPT-4.1-mini, DeepSeek-V3.2~\citep{deepseek-v3.2}, and Qwen3.5-397B-A17B~\citep{qwen3.5}. The results demonstrate that SafeRun achieves 100\% safety score (vs.\ 79.1\% PE average and 97.6\% CodeAct average) while maintaining competitive instruction-following scores.

In summary, our contributions are three-fold:
(1) A novel framework for deterministic LLM-based running planning, which decomposes determinism-critical planning into a soft interpretation phase and a hard enforcement phase.
(2) A comprehensive benchmark for running planning under realistic physiological and safety constraints.
(3) Extensive empirical evaluations, demonstrating that SafeRun consistently outperforms strong baselines in terms of both constraint satisfaction and planning quality across diverse scenarios.

\begin{figure*}[t]
\centering
\includegraphics[width=0.88\textwidth]{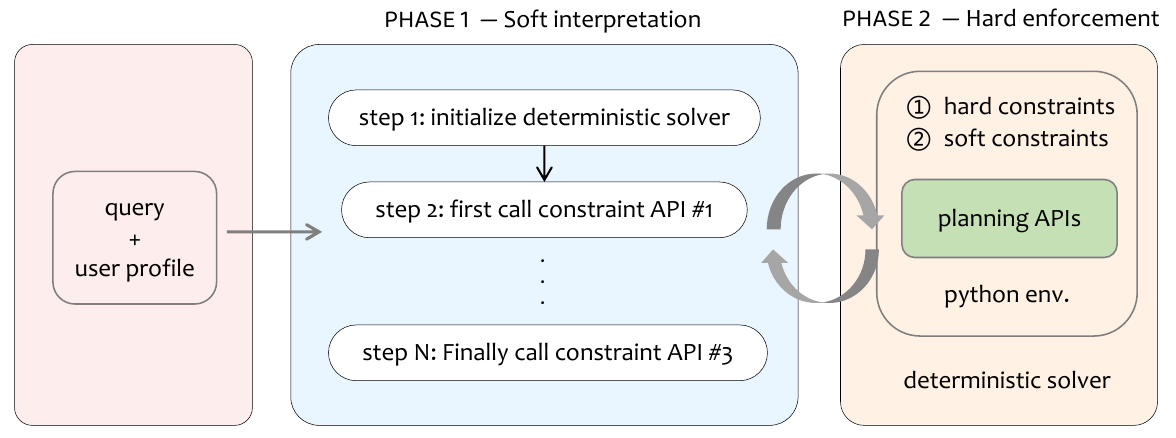}
\caption{Overview of SafeRun. Given a query and a user profile, SafeRun decouples planning into two stages. In \textbf{Phase~1 (soft interpretation)}, an LLM understands user intent, and construct calling pipeline, where it is restricted to invoking only predefined planning APIs. In \textbf{Phase~2 (hard enforcement)}, a deterministic solver executes the pipeline, where hard constraints (e.g., safety rules) and soft constraints (e.g., diversity and balance objectives) are enforced, yielding a running plan that is safety guaranteed. The two phases operate in an agent loop: when constraints cannot be satisfied, feedback is returned to the LLM for replanning until a valid solution is found or a maximum number of iterations is reached.}
\vspace{-1em}
\label{fig:method}
\end{figure*}

\section{Method}
\label{Method}
\subsection{Problem Formulation}
\label{Problem Formulation}
Given a user profile $p$ and a natural-language query $q$ that specifies the user's planning requirements, such as plan duration, target weekly mileage, run frequency, and day-level preferences, our goal is to generate a multi-week running plan $\pi$. 
Following the Daniels' taxonomy for endurance training~\cite{daniels2013daniels}, each day $d$ in the plan is assigned a running type $z_d \in {\texttt{REST}, \texttt{E}, \texttt{M}, \texttt{T}, \texttt{I}, \texttt{R}}$ and a distance $\ell_d \ge 0$. Here, \texttt{REST} denotes a rest day, while \texttt{E}, \texttt{M}, \texttt{T}, \texttt{I}, and \texttt{R} denote Easy, Marathon-pace, Threshold, Interval, and Repetition sessions, respectively.

To ensure training safety and physiological plausibility, the generated plan must satisfy a set of domain-level constraints regardless of how user requirements vary. In other words, while the system should adapt to personalized requests, all feasible outputs must remain within predefined safety boundaries. In this work, we adopt the following safety standards\footnote{The safety rules are derived from Daniels' coaching heuristics~\cite{daniels2013daniels} and may require domain-specific calibration.}.

\begin{itemize}
    \item \textbf{At most two intensity session per week.} Each training week may contain no more than two high-intensity workout, where intensity sessions include \texttt{T}, \texttt{I}, and \texttt{R}.
    \item \textbf{No consecutive intensity sessions.} High-intensity workouts must be separated by non-intensity day (i.e., \texttt{REST}, \texttt{E}, or \texttt{M}) for sufficient recovery.
    \item \textbf{Weekly mileage upper bound.} The total running distance within each week must not exceed a predefined maximum distance .
    \item \textbf{Per-session mileage upper bound.} The distance assigned to any single training day must not exceed $50\%$ of the total weekly mileage.
    \item \textbf{Intensity mileage ratio constraint.} The mileage of intensity sessions within a week must not exceed $10\%$ of the total weekly mileage.
\end{itemize}

Formally, let $\mathcal{W}$ denote the set of weeks in plan $\pi$, and let $\mathcal{D}(w)$ be the set of days in week $w$. Let the indicator function $\mathbb{I}(\cdot)$ equal $1$ if the condition is true and $0$ otherwise. Let $\mathcal{Z}_{int}=\{\texttt{T},\texttt{I},\texttt{R}\}$ denote the set of intensity workout types. A final plan must satisfy:
\begin{align}
\sum_{d \in \mathcal{D}(w)} \mathbb{I}(z_d \in \mathcal{Z}_{int}) &\le 2, \\
\mathbb{I}(z_d \in \mathcal{Z}_{int}) + \mathbb{I}(z_{d+1} \in \mathcal{Z}_{int}) &\le 1, \\
\sum_{d \in \mathcal{D}(w)} \ell_d &\le L^{\max}_{w}, \\
\ell_d &\le 0.5 \sum_{d \in \mathcal{D}(w)} \ell_d, \\
\sum_{d \in \mathcal{D}(w)} \ell_d \mathbb{I}(z_d \in \mathcal{Z}_{int})
&\le 0.1 \sum_{d \in \mathcal{D}(w)} \ell_d
\end{align}
where $L^{\max}_{w}$ denote the maximum allowed weekly mileage.

\subsection{SafeRun}

\paragraph{Overview.}

As illustrated in Figure~\ref{fig:method}, SafeRun follows a decoupled generate--execute CodeAct agent architecture. Given a query and a user profile, SafeRun decouples planning into two stages. In Phase~1 (soft interpretation), an LLM understands user intent, and construct calling pipeline, where it is restricted to invoking only predefined planning APIs. In Phase~2 (hard enforcement), a deterministic solver executes the pipeline, where hard constraints (e.g., safety rules) and soft constraints (e.g., diversity and balance objectives) are enforced, yielding a running plan that is safety guaranteed. The two phases operate in an agent loop: when constraints cannot be satisfied, feedback is returned to the LLM for replanning until a valid solution is found or a maximum number of iterations is reached. 

\paragraph{Deterministic Solver.} The deterministic solver in SafeRun is implemented as a CP-SAT based system composed of three tightly integrated components. First, hard constraints (e.g., safety rules) encode the safety rules in Section ~\ref{Problem Formulation}. Second, soft constraints (e.g., diversity and balance objectives) define optimization preferences that improve training quality, such as evenly distributing workload across the week, and avoiding clustering of similar workouts. Third, predefined planning APIs provide a structured interface for translating user requirements into solver inputs. For example, APIs such as $\texttt{set\_day\_type(day, running\_type)}$ allow the LLM to declare constraints explicitly without directly manipulating the optimization logic (Refer to Appendix~\ref{app:Implementation Details} for more details about soft constraints and planning APIs).  It is important to note that both hard constraints and soft objectives are explicitly hard-coded and strictly enforced within the solver.
Consequently, by combining these three components within a unified framework, the solver guarantees feasibility with respect to safety and user requirements while optimizing for high-quality running schedules. 

\paragraph{Benchmark Construction.} We construct a benchmark to evaluate both natural-language understanding and hard-constraint satisfaction under realistic user diversity through a multi-stage pipeline: (1) Query Generation: 10 human experts (bachelor's degree or higher) manually curate a large pool of candidate multi-week running planning queries covering diverse training goals, constraints, and user intentions; (2) Quality Filtering: the initial query pool is rigorously filtered to remove ambiguous or low-quality samples. To facilitate fair comparison across methods, we retain only data that conforms to safety rules, resulting in 100 high-quality and diverse natural-language queries; (3) Profile Construction: we construct 10 distinct runner profiles reflecting varying physiological conditions and training levels, including beginner, intermediate, and advanced runners with different mileage capacities and constraints; and (4) Randomly Pairing: each query is randomly paired with 4 user profile, yielding 400 profile--query combinations as the final evaluation benchmark.

\section{Experiments}

\begin{table*}[t]
\centering
\caption{Success rate, safety, and instruction-following performance across models and methods.}
\label{tab:main_results}
\resizebox{1.95\columnwidth}{!}{%
\small
\begin{tabular}{lccccccccc}
\toprule
\multirow{2}{*}{\textbf{Model}} 
& \multicolumn{3}{c}{\textbf{Success (\%)}} 
& \multicolumn{3}{c}{\textbf{Safety (\%)}} 
& \multicolumn{3}{c}{\textbf{Instruction (\%)}} \\
\cmidrule(lr){2-4} \cmidrule(lr){5-7} \cmidrule(lr){8-10}
& PE & CodeAct & SafeRun 
& PE & CodeAct & SafeRun 
& PE & CodeAct & SafeRun \\
\midrule
GPT-5 & 100.0 & 100.0 & 100.0 & 99.8 & 100.0 & \textbf{100.0} & 85.2 & 84.3 & \textbf{87.3} \\
GPT-4.1 & 100.0 & 100.0 & 100.0 & 89.5 & 96.0 & \textbf{100.0} & \textbf{81.5} & 78.7 & 79.6 \\
GPT-4.1-mini & 100.0 & 100.0 & 100.0 & 67.8 & 96.2 & \textbf{100.0} & 54.4 & 76.4 & \textbf{79.9} \\
DeepSeek-V3.2 & 100.0 & 100.0 & 100.0 & 76.5 & 96.5 & \textbf{100.0} & 66.2 & \textbf{79.5} & 78.7 \\
Qwen3.5-397B-A17B & 100.0 & 100.0 & 100.0 & 62.0 & 99.5 & \textbf{100.0} & 58.8 & \textbf{82.4} & 82.1 \\
\midrule
Average & 100.0 & 100.0 & 100.0 & 79.1 & 97.6 & \textbf{100.0} & 69.2 & 80.3 & \textbf{81.5} \\
\bottomrule
\end{tabular}
}
\vspace{-1em}
\end{table*}

\subsection{Experimental Setup}

\paragraph{Implementation Details.} We implement SafeRun using OR-Tools \cite{ortools} for constraint programming and smolagents \cite{smolagents} to provide a sandboxed Python execution environment for code execution. For all experiments, we set a maximum of 5 iterations for agent loop. Refer to Appendix~\ref{app:Implementation Details} for more details of implementation, such as soft constraints and planning APIs.

\paragraph{Evaluated LLMs and Baselines.} We evaluate five state-of-the-art proprietary and open-source LLMs, including GPT-5~\citep{gpt5}, GPT-4.1~\citep{gpt4.1}, GPT-4.1-mini, DeepSeek-V3.2~\citep{deepseek-v3.2}, and Qwen3.5-397B-A17B~\citep{qwen3.5}. We compare SafeRun against two representative baselines: (1) prompt engineering (PE), where the LLM is instructed to directly generate a valid training plan with all safety constraints explicitly specified in the system prompt; and (2) CodeAct, which follows prior tool-integrated approaches~\cite{hao2025large} by allowing the LLM to generate and execute general-purpose code for planning.

\paragraph{Evaluation Metrics.} We report three complementary metrics: 
(1)~\textbf{Success Score}, defined as the proportion of generated a valid plan in given iterations; (2)~\textbf{Safety Score}, the proportion of generated plans that satisfy all safety rules under rule-based verification; (3)~\textbf{Instruction-Following Score}, measured using an LLM-as-a-Judge (Claude Opus 4.6, refer to Appendix~\ref{app:judge-prompt} for evaluation prompt) that scores each plan on a 0--100 scale assessing how well it adheres to user-specified requirements such as duration, weekly volume, run frequency, and intensity placement.

\subsection{Main Results}

\paragraph{Success Score.} As shown in in Table~\ref{tab:main_results}, thanks to the curation process of benchmark: human experts strictly filtered the dataset to ensure every query-profile combination has at least one feasible solution, all evaluated methods achieve a \textbf{100.0\% success score} across all evaluated models.

\paragraph{Safety Score.} SafeRun achieves \textbf{100.0\% safety score} across all evaluated models, significantly outperforming prompt engineering (79.1\% on average) and CodeAct (97.6\%). This demonstrates that the deterministic solver provides strict and consistent enforcement of all safety constraints. Even models with lower baseline safety, such as GPT-4.1-mini (67.8\% $\to$ 100.0\%) and Qwen3.5-397B-A17B (62.0\% $\to$ 100.0\%), achieve full compliance under SafeRun, which is critical for reliable deployment.

\paragraph{Instruction-Following Score.} SafeRun achieves an average score of \textbf{81.5\%}, outperforming prompt engineering (69.2\%) and slightly exceeding CodeAct (80.3\%). The improvement is more evident on weaker models, while performance remains stable for stronger models such as GPT-5 and GPT-4.1, indicating that SafeRun improves alignment with user requirements without sacrificing robustness.

\subsection{More Analysis}

\paragraph{The effect of deterministic solver.} As shown in Table~\ref{tab:main_results}, compared with CodeAct, which is given a Python tool without deterministic solver, SafeRun obtain a consistent 100.0\% safety score across all models, suggesting that deterministic solver is the key factor behind SafeRun’s reliable constraint enforcement. We provide a  complete trace illustrating the progress of SafeRun in Appendix~\ref{app:cases}.

\paragraph{Benchmark analysis.} The results in Table~\ref{tab:main_results} show a consistent correlation between model capability and benchmark performance. Under prompt engineering, all models achieve 100.0\% success, while stronger models obtain higher safety and instruction-following scores, ranging from 62.0\% to 99.8\% in safety and from 58.8\% to 85.2\% in instruction following. A similar trend remains under SafeRun, where all models maintain 100.0\% success and safety, while instruction-following improves from 82.1\% to 87.3\% as model capability increases. These results indicate that the benchmark effectively captures underlying model differences while remaining sensitive to improvements from stronger planning frameworks.

\section{Limitation} Despite its effectiveness, SafeRun has several limitations. First, to ensure fair comparison across methods, our benchmark only includes feasible query–profile pairs, and thus does not evaluate infeasible or conflicting requests. Second, the deterministic solver relies on manually designed domain-specific rules and APIs, which may limit its generalization to other planning domains beyond running.

\section{Conclusion}
This paper proposes SafeRun, a framework for deterministic LLM-based planning via a decoupled architecture. SafeRun separates soft interpretation by an LLM from hard constraint enforcement by a deterministic solver, ensuring strict safety constraints while preserving natural-language flexibility. 

\section*{Impact Statement}
This work improves the reliability of LLM-based planning in safety-critical settings by introducing deterministic constraint enforcement. While demonstrated on running plans, the approach is general and applicable to other domains requiring strict constraints. SafeRun is intended as a decision-support tool and should not replace professional guidance. We emphasize responsible deployment when using LLM-based systems in safety-sensitive applications.

\nocite{langley00}

\bibliography{example_paper}

\begin{thebibliography}{24}
\providecommand{\natexlab}[1]{#1}
\providecommand{\url}[1]{\texttt{#1}}
\expandafter\ifx\csname urlstyle\endcsname\relax
  \providecommand{\doi}[1]{doi: #1}\else
  \providecommand{\doi}{doi: \begingroup \urlstyle{rm}\Url}\fi

\bibitem[Besta et~al.(2024)Besta, Blach, Kubicek, Gerstenberger, Podstawski, Gianinazzi, Gajda, Lehmann, Niewiadomski, Nyczyk, et~al.]{besta2024graph}
Besta, M., Blach, N., Kubicek, A., Gerstenberger, R., Podstawski, M., Gianinazzi, L., Gajda, J., Lehmann, T., Niewiadomski, H., Nyczyk, P., et~al.
\newblock Graph of thoughts: Solving elaborate problems with large language models.
\newblock In \emph{Proceedings of the AAAI Conference on Artificial Intelligence}, volume~38, pp.\  17682--17690, 2024.

\bibitem[Chen et~al.(2025)Chen, Tian, Ma, Xie, Chen, and Zhu]{chen2025unbiased}
Chen, M., Tian, J., Ma, L., Xie, D., Chen, W., and Zhu, J.
\newblock Unbiased evaluation of large language models from a causal perspective.
\newblock \emph{arXiv preprint arXiv:2502.06655}, 2025.

\bibitem[Chen et~al.(2023{\natexlab{a}})Chen, Arkin, Zhang, Roy, and Fan]{chen2023autotamp}
Chen, Y., Arkin, J., Zhang, Y., Roy, N., and Fan, C.
\newblock Autotamp: Autoregressive task and motion planning with llms as translators and checkers.
\newblock \emph{arXiv preprint arXiv:2306.06531}, 2023{\natexlab{a}}.

\bibitem[Chen et~al.(2023{\natexlab{b}})Chen, Arkin, Zhang, Roy, and Fan]{chen2023scalable}
Chen, Y., Arkin, J., Zhang, Y., Roy, N., and Fan, C.
\newblock Scalable multi-robot collaboration with large language models: Centralized or decentralized systems?
\newblock \emph{arXiv preprint arXiv:2309.15943}, 2023{\natexlab{b}}.

\bibitem[Daniels(2013)]{daniels2013daniels}
Daniels, J.
\newblock \emph{Daniels' running formula}.
\newblock Human Kinetics, 2013.

\bibitem[{Google}()]{ortools}
{Google}.
\newblock Or-tools.
\newblock URL \url{https://developers.google.com/optimization/}.
\newblock Accessed: 2026-03-01.

\bibitem[Guan et~al.(2023)Guan, Valmeekam, Sreedharan, and Kambhampati]{guan2023leveraging}
Guan, L., Valmeekam, K., Sreedharan, S., and Kambhampati, S.
\newblock Leveraging pre-trained large language models to construct and utilize world models for model-based task planning.
\newblock \emph{Advances in Neural Information Processing Systems}, 36:\penalty0 79081--79094, 2023.

\bibitem[Hao et~al.(2023)Hao, Gu, Ma, Hong, Wang, Wang, and Hu]{hao2023reasoning}
Hao, S., Gu, Y., Ma, H., Hong, J.~J., Wang, Z., Wang, D.~Z., and Hu, Z.
\newblock Reasoning with language model is planning with world model.
\newblock \emph{arXiv preprint arXiv:2305.14992}, 2023.

\bibitem[Hao et~al.(2025)Hao, Chen, Zhang, and Fan]{hao2025large}
Hao, Y., Chen, Y., Zhang, Y., and Fan, C.
\newblock Large language models can solve real-world planning rigorously with formal verification tools.
\newblock In \emph{Proceedings of the 2025 Conference of the Nations of the Americas Chapter of the Association for Computational Linguistics: Human Language Technologies (Volume 1: Long Papers)}, pp.\  3434--3483, 2025.

\bibitem[Kojima et~al.(2022)Kojima, Gu, Reid, Matsuo, and Iwasawa]{kojima2022large}
Kojima, T., Gu, S.~S., Reid, M., Matsuo, Y., and Iwasawa, Y.
\newblock Large language models are zero-shot reasoners.
\newblock \emph{Advances in neural information processing systems}, 35:\penalty0 22199--22213, 2022.

\bibitem[Liu et~al.(2025)Liu, Mei, Lin, Xue, Wang, Xu, Wu, Zhang, Lin, Dong, et~al.]{deepseek-v3.2}
Liu, A., Mei, A., Lin, B., Xue, B., Wang, B., Xu, B., Wu, B., Zhang, B., Lin, C., Dong, C., et~al.
\newblock Deepseek-v3. 2: Pushing the frontier of open large language models.
\newblock \emph{arXiv preprint arXiv:2512.02556}, 2025.

\bibitem[Liu et~al.(2023)Liu, Jiang, Zhang, Liu, Zhang, Biswas, and Stone]{liu2023llm+}
Liu, B., Jiang, Y., Zhang, X., Liu, Q., Zhang, S., Biswas, J., and Stone, P.
\newblock Llm+ p: Empowering large language models with optimal planning proficiency.
\newblock \emph{arXiv preprint arXiv:2304.11477}, 2023.

\bibitem[Madaan et~al.(2024)Madaan, Tandon, Gupta, Hallinan, Gao, Wiegreffe, Alon, Dziri, Prabhumoye, Yang, et~al.]{madaan2024self}
Madaan, A., Tandon, N., Gupta, P., Hallinan, S., Gao, L., Wiegreffe, S., Alon, U., Dziri, N., Prabhumoye, S., Yang, Y., et~al.
\newblock Self-refine: Iterative refinement with self-feedback.
\newblock \emph{Advances in Neural Information Processing Systems}, 36, 2024.

\bibitem[OpenAI(2025{\natexlab{a}})]{gpt4.1}
OpenAI.
\newblock Introducing {GPT-4.1} in the {API}, April 2025{\natexlab{a}}.
\newblock URL \url{https://openai.com/index/gpt-4-1/}.

\bibitem[OpenAI(2025{\natexlab{b}})]{gpt5}
OpenAI.
\newblock {GPT-5} system card, August 2025{\natexlab{b}}.
\newblock URL \url{https://arxiv.org/abs/2601.03267}.

\bibitem[{Qwen Team}(2026)]{qwen3.5}
{Qwen Team}.
\newblock {Qwen3.5}: Towards native multimodal agents, February 2026.
\newblock URL \url{https://qwen.ai/blog?id=qwen3.5}.

\bibitem[Roucher et~al.(2025)Roucher, del Moral, Wolf, von Werra, and Kaunismäki]{smolagents}
Roucher, A., del Moral, A.~V., Wolf, T., von Werra, L., and Kaunismäki, E.
\newblock `smolagents`: a smol library to build great agentic systems.
\newblock \url{https://github.com/huggingface/smolagents}, 2025.

\bibitem[Shen et~al.(2024)Shen, Song, Tan, Li, Lu, and Zhuang]{shen2024hugginggpt}
Shen, Y., Song, K., Tan, X., Li, D., Lu, W., and Zhuang, Y.
\newblock Hugginggpt: Solving ai tasks with chatgpt and its friends in hugging face.
\newblock \emph{Advances in Neural Information Processing Systems}, 36, 2024.

\bibitem[Shinn et~al.(2024)Shinn, Cassano, Gopinath, Narasimhan, and Yao]{shinn2024reflexion}
Shinn, N., Cassano, F., Gopinath, A., Narasimhan, K., and Yao, S.
\newblock Reflexion: Language agents with verbal reinforcement learning.
\newblock \emph{Advances in Neural Information Processing Systems}, 36, 2024.

\bibitem[Wang et~al.(2024)Wang, Chen, Yuan, Zhang, Li, Peng, and Ji]{wang2024executable}
Wang, X., Chen, Y., Yuan, L., Zhang, Y., Li, Y., Peng, H., and Ji, H.
\newblock Executable code actions elicit better llm agents.
\newblock In \emph{Forty-first International Conference on Machine Learning}, 2024.

\bibitem[Wei et~al.(2022)Wei, Wang, Schuurmans, Bosma, Xia, Chi, Le, Zhou, et~al.]{wei2022chain}
Wei, J., Wang, X., Schuurmans, D., Bosma, M., Xia, F., Chi, E., Le, Q.~V., Zhou, D., et~al.
\newblock Chain-of-thought prompting elicits reasoning in large language models.
\newblock \emph{Advances in neural information processing systems}, 35:\penalty0 24824--24837, 2022.

\bibitem[Yao et~al.(2022)Yao, Zhao, Yu, Du, Shafran, Narasimhan, and Cao]{yao2022react}
Yao, S., Zhao, J., Yu, D., Du, N., Shafran, I., Narasimhan, K., and Cao, Y.
\newblock React: Synergizing reasoning and acting in language models.
\newblock \emph{arXiv preprint arXiv:2210.03629}, 2022.

\bibitem[Yao et~al.(2024)Yao, Yu, Zhao, Shafran, Griffiths, Cao, and Narasimhan]{yao2024tree}
Yao, S., Yu, D., Zhao, J., Shafran, I., Griffiths, T., Cao, Y., and Narasimhan, K.
\newblock Tree of thoughts: Deliberate problem solving with large language models.
\newblock \emph{Advances in Neural Information Processing Systems}, 36, 2024.

\bibitem[Zhao et~al.(2024)Zhao, Lee, and Hsu]{zhao2024large}
Zhao, Z., Lee, W.~S., and Hsu, D.
\newblock Large language models as commonsense knowledge for large-scale task planning.
\newblock \emph{Advances in Neural Information Processing Systems}, 36, 2024.

\end{thebibliography}
\bibliographystyle{icml2026}

\newpage
\appendix
\onecolumn
\appendix

\section{Related Works}
\label{Related Works}
\noindent \textbf{LLM Planning.} 
LLMs have demonstrated strong potential in reasoning~\cite{wei2022chain, kojima2022large, yao2022react}. Existing approaches mainly improve planning through several directions: pure LLM based methods~\cite{wei2022chain, yao2022react, shen2024hugginggpt, shinn2024reflexion, madaan2024self, chen2023scalable}, and integrated with external planners~\cite{liu2023llm+, guan2023leveraging, chen2023autotamp, hao2025large}. 
However, our method differs from prior work in two key aspects. First, existing studies mainly focus on general planning scenarios, where failures are inconvenient but usually non-critical, whereas we target determinism-critical domains such as running planning, where violating physiological constraints may cause injury. Second, previous LLM-planning approaches typically rely on unconstrained planners. In contrast, our framework combines LLM flexibility with deterministic constraint enforcement.

\noindent \textbf{ReAct and CodeAct.}
ReAct~\cite{yao2022react} combines reasoning and acting by interleaving natural-language thoughts with external actions, enabling LLMs to solve tasks through iterative interaction with tools or environments. CodeAct\cite{wang2024executable} extends this paradigm by allowing models to generate executable code as actions. 
Built upon CodeAct, SafeRun separates language understanding from constraint enforcement by restricting the LLM to invoke a predefined deterministic solver, ensuring every generated plan satisfies all safety requirements.

\section{Implementation Details}
\label{app:Implementation Details}
\subsection{Soft Constraints}
Beyond hard safety rules, the deterministic solver optimizes for running plan quality through three soft constraints.
\begin{itemize}
    \item Workout diversity: random perturbations to avoid repetitive schedules
    \item Run/rest balance: penalizing consecutive running days (weight -5) and 
consecutive rest days (weight -3) to promote alternating patterns
    \item Distance uniformity: minimizing variance in daily mileage across running 
days to ensure balanced workload distribution within each week.
\end{itemize}

\subsection{planning APIs}
The core APIs include:

\begin{itemize}
    \item \textbf{Initialization}: \texttt{Planner} sets global training parameters (weekly volume, run frequency, plan duration).
    
    \item \textbf{Assign workout types}: \texttt{set\_day\_type} assigns specific workout types (e.g., day 3 = Interval run).
    
    \item \textbf{Constrain distances}: \texttt{set\_day\_distance\_range} bounds daily mileage (e.g., day 5 between 10--15 km).
    

    \item \textbf{Execute and display}: \texttt{solve\_and\_print} executes the solver and prints the formatted training plan.
    
\end{itemize}

\section{Prompts}
\label{app:prompts}

\subsection{SafeRun Prompt}
\label{app:codeact-system-prompt}

\begin{tcolorbox}[colback=gray!5, colframe=gray!60, title=SafeRun Prompt, breakable, fonttitle=\bfseries\small, fontupper=\scriptsize]
\begin{verbatim}
# Role: Professional Running Coach

You are a professional running coach. Your task is to generate safe, personalized training plans.

## Pace Zones
- **E (Easy)**: Recovery and base building
- **M (Marathon Pace)**: Aerobic endurance
- **T (Threshold)**: Lactate threshold training
- **I (Interval)**: High-intensity intervals
- **R (Repetition)**: Short sprints for speed

## Input Data
You will receive:
1. **User profile**: age, gender, height(cm), weight(kg), pace_zones (E/M/T/I/R)
2. User query with constraints (weekly volume, runs per week, specific day requirements)

## Hard Constraints (System Enforced)
- Supports **multi-week planning** 
- **Safety first**: If query is unsafe, warn and suggest alternative

## Planning Guidelines

Based on user profile (age, gender, BMI from height/weight):
- **Younger/Lighter**: Can handle higher intensity
- **Older/Heavier**: Focus on base building, lower intensity
- **BMI**: Calculate from height(cm) and weight(kg) to estimate fitness level

## Query Constraint Extraction

**CRITICAL: Parse these from user query before generating code:**

1. **num_weeks**: Look for "X weeks" or "X-week"
2. **weekly_km**: Look for "X km", "X km/week"
3. **runs_per_week**: Look for "X runs", "X sessions"
4. **day constraints**: Map to absolute days (day 1 = Monday of Week 1)

## Workflow

**If user query is valid:**
1. **Extract** all constraints from query (num_weeks, weekly_km, runs_per_week, day constraints)
2. Call `python_executor` with `RunningSchedulePlanner`
3. Set `weekly_km`, `runs_per_week`, and `num_weeks` to match query exactly
4. Add user constraints
5. Output plan
\end{verbatim}
\end{tcolorbox}

\subsection{SafeRun Tool Description}
\label{app:codeact-tool-prompt}

\begin{tcolorbox}[colback=blue!3, colframe=blue!40, title=SafeRun Tool Description, breakable, fonttitle=\bfseries\small, fontupper=\scriptsize]
\begin{verbatim}
## Python Code Executor

**Pre-loaded Classes (DO NOT import, use directly):**
- `RunningSchedulePlanner` - for single week generation
- `SequentialPlanner` - for multi-week week-by-week generation

### For Multi-Week Plans (Recommended for Precise Control)

Use `SequentialPlanner` to generate week by week. This ensures EXACT weekly volume and run frequency per week:

```python
# SequentialPlanner is pre-loaded, use directly
planner = SequentialPlanner(pace_zones={'E': '5:56', 'M': '5:29', 'T': '5:06'})

# Create weekly params - each week has exact specifications
weekly_params = [
    {"weekly_km": 40, "runs_per_week": 4, "constraints": [{"day": 3, "zone": "I"}]},
    {"weekly_km": 40, "runs_per_week": 4, "constraints": [{"day": 3, "zone": "I"}]},
    {"weekly_km": 40, "runs_per_week": 4, "constraints": [{"day": 3, "zone": "I"}]},
    {"weekly_km": 40, "runs_per_week": 4, "constraints": [{"day": 3, "zone": "I"}]},
]

plan = planner.generate_multi_week(num_weeks=4, weekly_params=weekly_params)
import json
print(json.dumps(plan, indent=2, ensure_ascii=False))
```

### For Single Week

Use `RunningSchedulePlanner` directly:

```python
# RunningSchedulePlanner is pre-loaded, use directly
planner = RunningSchedulePlanner(
    weekly_km=50.0,
    pace_zones={'E': '5:56', 'M': '5:29'},
    runs_per_week=5,
    num_weeks=1
)
planner.set_day_type(3, zone='I')  # Wednesday interval
planner.solve_and_print()
```

**API:**
- `SequentialPlanner(pace_zones)` - for multi-week sequential generation
  - `generate_multi_week(num_weeks, weekly_params)` - returns combined plan
- `RunningSchedulePlanner(weekly_km, pace_zones, runs_per_week, num_weeks=1)`
  - `set_day_type(day, zone)` - zone: REST/E/M/T/I/R, day: 1-7
  - `set_day_distance_range(day, min_km=None, max_km=None)` - constrain distance for a specific day, day: 1-7
  - `solve_and_print()` - solves and prints the plan
  - `solve()` - solves and returns plan as dict (no printing)

**Built-in Hard Constraints (auto-enforced):**
- At most 1 intensity run (T/I/R) per week
- No consecutive intensity runs
- Intensity run <= 10% of weekly volume
- Single run <= 50% of weekly volume
- Weekly total <= 100 km (safety cap)
\end{verbatim}
\end{tcolorbox}

\subsection{LLM-as-a-Judge Prompt}
\label{app:judge-prompt}

\begin{tcolorbox}[colback=orange!3, colframe=orange!40, title=LLM-as-a-Judge Evaluation Prompt, breakable, fonttitle=\bfseries\small, fontupper=\scriptsize]
\begin{verbatim}
As a running training plan evaluation expert, please assess
whether the following training plan meets the user's query
requirements.

User Query: {query}

Weekly Statistics (pre-computed):
{weekly_stats}

Complete Plan: {plan_json}

Please check if the plan satisfies the following requirements
from the query:
1. Plan duration (number of weeks)
2. Weekly volume requirement - refer to the km values in
   "Weekly Statistics" above
3. Weekly training frequency - refer to the runs count in
   "Weekly Statistics" above
4. Specific rest day/intensity day arrangements
5. Other explicit constraints

Please respond in JSON format:
{
    "satisfied": true/false,
    "score": 0-100,
    "reasoning": "brief explanation"
}
\end{verbatim}
\end{tcolorbox}

\section{Case Studies}
\label{app:cases}

We present a complete case for both the SafeRun and Baseline approaches on the same query.

\vspace{0.5em}
\noindent\textbf{Query (sample\_0227):} \textit{``My 10K is in two months---please deliver an 8-week plan at 50\,km per week with 5 runs and one T session each week.''}

\noindent\textbf{User Profile:} Age 20, Male, 180\,cm, 64\,kg, Pace Zones: E\,=\,4:24, M\,=\,4:02, T\,=\,3:43, I\,=\,3:33, R\,=\,3:33

\subsection{SafeRun}
\label{app:case-codeact}

\paragraph{Step 1: LLM receives system prompt + tool description + user query + user profile.}

The model is provided with the system prompt (Appendix~\ref{app:codeact-system-prompt}), the tool description (Appendix~\ref{app:codeact-tool-prompt}), and the following user message:

\begin{tcolorbox}[colback=gray!5, colframe=gray!50, title=User Message, fonttitle=\bfseries\small, fontupper=\scriptsize]
\begin{verbatim}
User Profile:
- Age: 20, Gender: male, Height: 180cm, Weight: 64kg
- Pace Zones: {'E':'4:24','M':'4:02','T':'3:43',
               'I':'3:33','R':'3:33'}

User Query:
My 10K is in two months—please deliver an 8-week plan
at 50 km per week with 5 runs and one T session each week.

IMPORTANT: Use SequentialPlanner. Extract num_weeks, weekly_km,
runs_per_week, and intensity/rest constraints from the query.
\end{verbatim}
\end{tcolorbox}

\paragraph{Step 2: LLM generates a tool call.}

The model emits a \texttt{python\_executor} function call with the following code:

\begin{tcolorbox}[colback=green!3, colframe=green!50, title=Tool Call: \texttt{python\_executor}, fonttitle=\bfseries\small, fontupper=\scriptsize]
\begin{verbatim}
# ===== Extracted from query =====
# num_weeks = 8
# weekly_km = 50
# runs_per_week = 5
# constraints: one T session each week (no specific day)
# ================================

import json

planner = SequentialPlanner(
    pace_zones={'E':'4:24','M':'4:02',
                'T':'3:43','I':'3:33','R':'3:33'})

weekly_params = []
for week in range(8):
    weekly_params.append({
        "weekly_km": 50,
        "runs_per_week": 5,
        "constraints": [{"zone": "T"}]
    })

plan = planner.generate_multi_week(
    num_weeks=8, weekly_params=weekly_params)
print(json.dumps(plan, indent=2, ensure_ascii=False))
\end{verbatim}
\end{tcolorbox}

\paragraph{Step 3: Tool execution.}

The codes are executed in the sandboxed Python
execution environment, enforcing all hard and soft constraints. We show Week~1 and Week~2 as examples (all 8 weeks follow the same pattern):

\begin{tcolorbox}[colback=yellow!5, colframe=yellow!50, title=Planning Results of SafeRun (Weeks 1--2 of 8), fonttitle=\bfseries\small, fontupper=\scriptsize]
\begin{verbatim}
W1-Monday:    Easy      11.9 km    W2-Monday:    Easy      10.9 km
W1-Tuesday:   REST       0.0 km    W2-Tuesday:   Easy      12.9 km
W1-Wednesday: Threshold  5.0 km    W2-Wednesday: Threshold   5.0 km
W1-Thursday:  Easy      11.9 km    W2-Thursday:  REST        0.0 km
W1-Friday:    Easy      10.3 km    W2-Friday:    Easy       10.9 km
W1-Saturday:  REST       0.0 km    W2-Saturday:  REST        0.0 km
W1-Sunday:    Easy      10.9 km    W2-Sunday:    Easy       10.3 km
         Week 1: 50.0 km                   Week 2: 50.0 km
\end{verbatim}
\end{tcolorbox}

Safety guaranteed by the solver across all 8~weeks:
\begin{itemize}[nosep]
    \item Weekly volume $=$ 50.0\,km exactly
    \item 5 runs + 2 rest days per week
    \item Exactly 1 T session per week, with T distance $\leq$ 5.0\,km $=$ 10\% of 50\,km
    \item No consecutive intensity runs; no intensity on Sunday followed by Monday
\end{itemize}

\paragraph{Step 4: Evaluation.}

\begin{itemize}[nosep]
    \item \textbf{Safety Score:} \textcolor{green!60!black}{\textbf{PASS}} --- 0 violations across all 8 weeks.
    \item \textbf{Instruction-Following Score:} \textbf{95/100} --- ``The plan meets all explicit requirements: (1)~8-week duration. (2)~Weekly volume of 50\,km for all 8 weeks. (3)~5 runs per week. (4)~Exactly one T session per week. Minor deduction: no taper in the final week before race day, but this was not explicitly requested.''
\end{itemize}

\subsection{PE}
\label{app:case-baseline}

\paragraph{Step 1: LLM receives system prompt + user query  + user profile.}

The model receives the system prompt and the same user profile and query.

\paragraph{Step 2: LLM directly planning.}

Without tool calling, the model directly planning. We show the first two weeks:

\begin{tcolorbox}[colback=red!3, colframe=red!30, title=Planning Results of PE (Weeks 1--2 of 8), fonttitle=\bfseries\small, fontupper=\scriptsize]
\begin{verbatim}
W1-Monday:    Easy  10 km     W2-Monday:    Easy  10 km
W1-Tuesday:   REST   0 km     W2-Tuesday:   REST   0 km
W1-Wednesday: Easy  10 km     W2-Wednesday: M     10 km
W1-Thursday:  REST   0 km     W2-Thursday:  REST   0 km
W1-Friday:    T      5 km     W2-Friday:    T      5 km
W1-Saturday:  REST   0 km     W2-Saturday:  REST   0 km
W1-Sunday:    Easy  10 km     W2-Sunday:    Easy  10 km
         Week 1: 35 km               Week 2: 35 km
\end{verbatim}
\end{tcolorbox}

\paragraph{Step 3: Evaluation.}

The Baseline plan exhibits \textbf{three categories of failures}:

\begin{itemize}[nosep]
    \item \textbf{Safety Score:} \textcolor{red!70!black}{\textbf{FAIL}} --- \textbf{8 violations}. Every week violates the ``intensity run $\leq$ 10\% of weekly volume'' constraint: T session = 5\,km but weekly total = 35\,km, so intensity ratio = 14.3\% $>$ 10\%.
    \item \textbf{Instruction-Following Score:} \textbf{20/100} --- ``The plan fails on two critical requirements: (1)~Weekly volume is only 35\,km, 30\% below the 50\,km target. (2)~Only 4 runs per week instead of 5. The plan does correctly provide 8 weeks and one T per week, but the two most fundamental constraints are not met.''
\end{itemize}


\end{document}